# A Machine Learning-Based Framework for Discovering Huntington's Disease Stages: Integrating Graph Representation Learning and clustering to Uncover Progression Dynamics in Longitudinal Enroll-HD Dataset

LUBNA M. ABU ZOHAIR

School of Mathematical and Computer Sciences, Heriot-Watt University, Dubai, United Arab Emirates

la2015@hw.ac.uk

MARTA VALLEJO

School of Mathematical and Computer Sciences, Heriot-Watt University, Edinburgh, United Kingdom

MD Azher UDDIN

School of Mathematical and Computer Sciences, Heriot-Watt University, Dubai, United Arab Emirates

JOHN R. WOODWARD

School of Mathematical and Computer Sciences, Heriot-Watt University, Dubai, United Arab Emirates

HIND ZANTOUT

School of Mathematical and Computer Sciences, Heriot-Watt University, Dubai, United Arab Emirates

Huntington's disease (HD) is a progressive brain disorder that gradually affects movement, cognitive function, and behavior. Identifying the stage of the disease accurately and consistently is important for understanding its course, grouping patients, personalized care, and discovering treatment. Existing clinical staging frameworks rely primarily on predefined clinical measurement thresholds and clinical expert decisions, yet these discrete cut-offs may obscure meaningful intra-stage variability and remain vulnerable to inter-rater differences, especially in motor and functional assessments. To address these limitations, we developed an unsupervised machine learning framework that leverages graph-based representation learning to capture the temporal relationships within and between patients. This approach encodes longitudinal clinical data into compact latent representations that reflect evolving disease patterns without relying on established disease staging clinical measurements. Using these learned representations, we applied K-means++ clustering to identify the number of well-separated groups. Then, Number of clusters, $k$ was iteratively incremented, while assessing defined clusters robustness through stability analysis. This approach resulted in the discovery of additional meaningful disease stages that preserved robustness and statistically significant distinctions, as assessed by PERMANOVA. We applied this framework to 302 individuals from the Enroll-HD cohort, encompassing 1,477 visits and 44 clinical variables per visit, of which 80% were manifest participants, enabling a data-driven discovery of HD stages grounded in the natural progression of clinical features. Despite the limited cohort size, the proposed framework achieved robust clustering performance, with a Silhouette score of 0.67, a Davies–Bouldin index of 0.56, and a Calinski–Harabasz score of 453, using a four-dimensional latent space that revealed two primary disease stages. The stability analysis also revealed four meaningful and significantly different stages, each corresponding to clearly defined clinical measurement boundaries, demonstrating a clear alignment with minimal overlap between stages compared to previously established clinical staging methods. This framework provides an objective, data-driven foundation for HD staging and has potential to integrate objective multimodal biomarkers to generate meaningful, objective insights into the course of HD disease.

## 1 INTRODUCTION

Understanding existing courses of a disease is particularly challenging in neurodegenerative diseases (NDDs), such as Huntington's disease (HD), as they exhibit abnormal dynamic changes in clinical symptoms and biomarkers [1, 2]. HD is an autosomal dominant disorder caused by a cytosine–adenine–guanine (CAG) trinucleotide repeat expansion in the HTT gene, leading to progressive neurodegeneration involving motor, cognitive, and broader functional brain systems [3]. Furthermore, the differences in genetic expression across individuals contribute to variability in age of onset, patterns of symptom progression, and the rate of multidomain impairment, including motor, cognitive, and functional domains [3]. Such multidomain longitudinal heterogeneity makes it difficult to understand how the disease progresses, track clinical changes, and select the best treatment strategies.

Several clinical frameworks have been developed to characterize HD progression, reflecting the complexity of its motor, functional, and biological manifestations. Traditionally, clinical diagnosis of manifest HD relies on the Diagnostic Confidence Level (DCL) of the Unified HD Rating Scale (UHDRS), where manifest HD is defined as DCL=4, corresponding to 99% diagnostic confidence based on unequivocal motor abnormalities assessed by a clinical expert [4]. Additionally, a functional staging system is used to classify impairment in patients' daily living and occupational capacities into five stages using the Shoulson and Fahn Total Functional Capacity (TFC) scale within the UHDRS [5, 6]. HD Integrated Staging System (HD-ISS) was recently introduced as a biologically informed framework that integrates genetic expansion, imaging biomarkers (putamen and caudate volumes from MRI), clinical measures (UHDRS motor or cognitive scores), and functional decline to stage disease progression across premanifest and manifest HD over the lifespan, though not standardized yet [7, 8]. Despite these advances, these staging systems remain dependent on expert clinician assessment of patient status, as well as predefined cutoffs and consensus-derived criteria that define diagnostic and disease stage thresholds [7, 8]. For example, clinicians confident about the patient state is required, however, this introduces inter-rater variability [9, 10]. Transitions between HD-ISS stages are operationalized using quantitative thresholds in imaging and clinical assessments proposed by the HD-ISS consortium, distinguishing Stage 1 (biomarker evidence) from Stage 2 (clinical signs), and Stage 2 from Stage 3 (functional decline) [7, 8]. Similarly, the Shoulson and Fahn staging system classifies individuals with HD into stages using predefined TFC score ranges [5]. These limitations motivate complementary approaches that objectively discover progression patterns directly from longitudinal clinical measurements or biomarkers, enabling a more precise and unbiased understanding of disease progression without reliance on fixed categorical rules [2].

Unsupervised machine learning (ML) approaches offer a promising alternative, as they can learn and group temporal patterns of disease progression directly from longitudinal clinical data without relying on predefined stage cutoffs [11–13]. Among these, dynamic graph-based models provide a principled framework for capturing the evolving relationships among, cognitive, and functional features over time [14, 15]. In particular, the unsupervised spatial–temporal fused node embedding framework (URL-STFN) proposed by Abuzohair et al. [16] demonstrated improved performance motor in Parkinson's disease staging compared with conventional representation learning and clustering approaches. In this work, the URL-STFN framework is evaluated for HD staging and further extended to incorporate a clustering stability assessment (illustrated in Figure 1) to better uncover further underlying disease progression patterns, including transitional states and imbalanced group structures, that may not be adequately captured by optimal clustering performance alone. We hypothesize that this framework can identify clinically meaningful HD stages in a more data-driven manner, exhibiting reduced feature value dispersion overlap and more well-defined boundaries than conventional clinical staging systems. more well-defined boundaries than conventional clinical staging systems. This work provides proof of concept framework for HD staging; necessitate evaluating on larger cohorts and clinical expert evaluation to assess the practical relevance of the identified stages before translational use.

## 2 METHOD

### 2.1 Dataset and Preparation

Data used in this work were generously provided by the participants in the Enroll-HD study and made available by the CHDI Foundation, Inc. Enroll-HD is a global clinical research platform designed to facilitate clinical research in HD. Core datasets are collected annually from all research participants as part of this multi-center longitudinal observational study. Data are monitored for quality and accuracy using a risk-based monitoring approach. All sites are required to obtain and maintain local ethical approval. In particular, patients' longitudinal records in the Enroll table in Enroll-HD PDS7-R1 clinical dataset were mainly utilized in providing the concept of the proposed model. This table initially comprised around 30,511 unique participants and 129,537 recorded visits [17-19].

We limited the analysis to patients who had records across at least four consecutive annual clinical visits. This selection helps ensure that potential disease states or transitions are not missed due to gaps in yearly visits. By requiring shared longitudinal coverage across patients, all individuals are evaluated over comparable time points, reducing the bias caused by missing or irregular observations. For example, if Patient 1 has clinical visits recorded at ages 42, 43, 44, and 45, then Patient 2, Patient 3, . . . Patient n should also have visit records at these same ages (42–45). Only patients with observations at these common sequential ages are retained for the analysis.

The dataset preparation steps and preprocessing procedures involved the following steps:

1. Participants classified as family controls or genotype-negative were excluded. Filtering was performed using the common identified 'subjid' across the “enroll.csv” and “participant.csv” tables. Individuals were selected based on the 'hdcat_0' retaining only those with premanifest or manifest motor stages

2. A total of 44 motors, cognitive, and functional clinical measurements were selected, these were chosen because they are commonly utilized in prior clinical staging studies were selected [5, 9, 20], and enabling results discussion and interpretation of the discovered disease stages. These clinical measurements are (more description about the variables listed below is available in Enroll-HD data dictionary file [9]):
   A. Motor features: rigarml, prosupr, dysarth, dystrle, sacinitv, fingtapr, brady, dysttrnk, dystrue, chorbol, gait, tongue, fingtapl, luria, chortrnk, chorlle, chorlue, chorface, sacvelh, ocular, prosupl, dystlle, sacinith, rigarmr, chorrue, tandem, chorrle, sacvelv, ocular, dystlue, and retropls.
   B. Functional features: occupath, finances, carelevl, chores, adl, and indepscl.
   C. Cognitive Features: verflt05, sit1, sdmt1, swrt1, scnt1, verfct5, mmsetotal

   Further details on these clinical measurements are available in the Enroll-HD data dictionary [9].

3. Aggregated features, such as CAP score and composite variables derived from clinical assessments (specifically motscore, tfcscore, and diagconf), were excluded from direct input to the models used for representation learning or clustering. These features were instead considered when discussing the characteristics of the disease stages derived from the selected granular clinical measurements.
4. Handling Missing Data. For temporal records that have all features missing (NaN), they were dropped. If this is the case with few missing features only, missing values were explicitly imputed with a constant value (-1) [21]. All clinical variables used for clustering are non-negative ordinal measures; therefore, -1 lies outside the physiologically plausible range and serves as an explicit indicator of missingness rather than a valid measurement. This strategy preserves missingness information while avoiding imputation-induced bias, dilution, or instability that can arise from low-performing ML–based imputations.
5. Features Standardization. The Enroll-HD dataset is known to contain clinical features measured on heterogeneous scales and to exhibit skewed distributions and extreme values. To mitigate the variable scaling influence, StandardScaler function was applied to standardize features to zero mean and unit variance, ensuring that variables measured on different scales contribute comparably to the analysis [22, 23].

### 2.2 Dynamic Graph Structure

Patients' longitudinal records were represented as an age-based dynamic graph ($G$), where each graph snapshot corresponds to a specific age and nodes represent patients with node features encoding their clinical measurements at that age. A temporal sequence of graphs was constructed to capture the evolution of patient profiles (represented as nodes in graph $G$) across consecutive ages, from the first common age ($Age1$) to the last common age ($AgeT$ ), as illustrated in Figure 1 (left panels with grey backgrounds). Age was chosen as the grouping anchor not to claim the age as a key reason of abnormal evolution but because it is one of the key risk factors to ageing and disease progression in HD and other neurodegenerative disorders, with progression often analyzed from age at onset, or age-based reference point [20, 24, 25]. This formulation enables the discovery of distinct disease stages that patients may variably traverse despite belonging to similar age ranges.

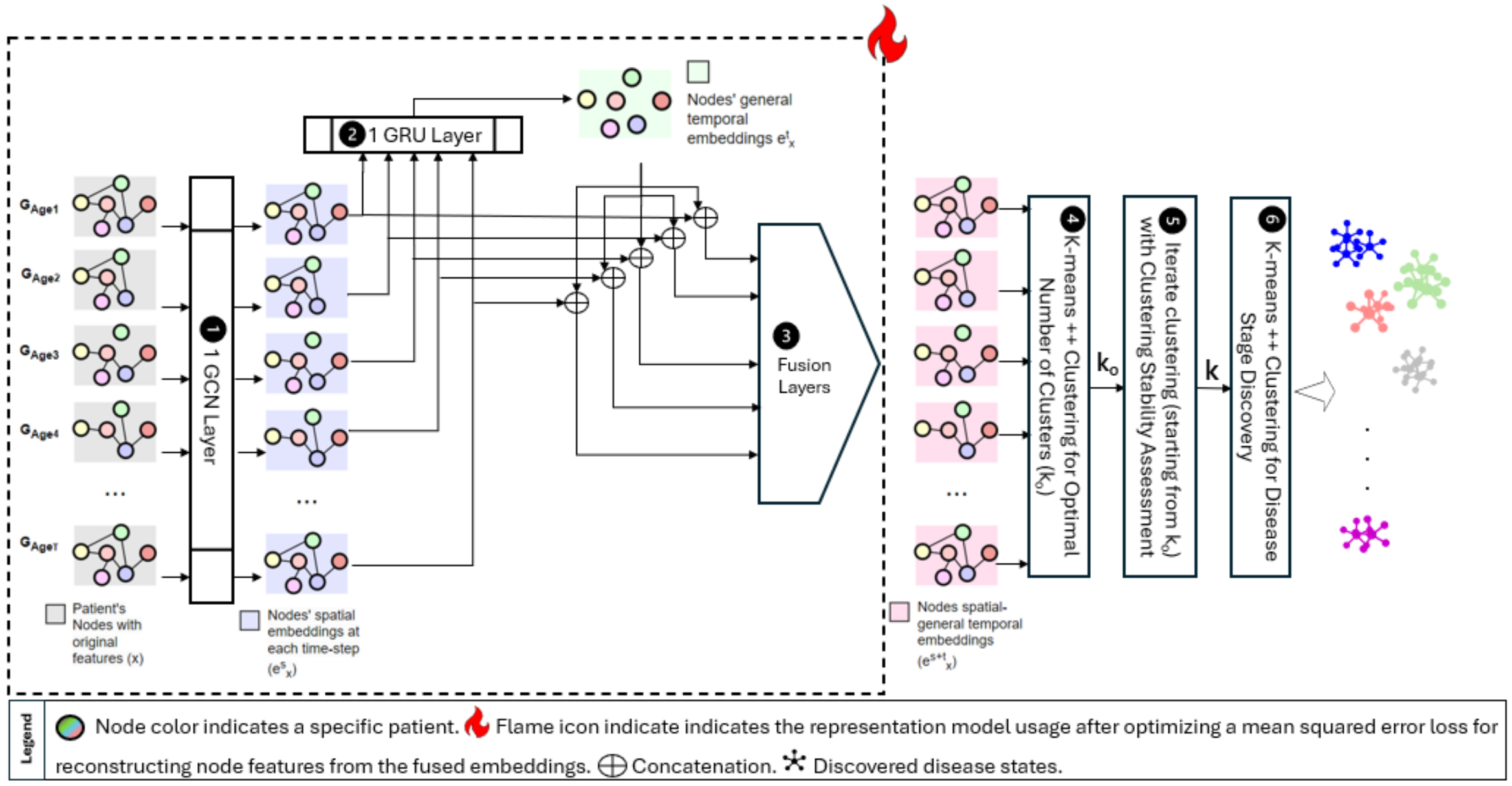

Figure 1: Framework architecture highlighting its key components, including the dynamic graph design and the representation model architecture, as proposed by Abuzohair et al. [2]

Edges within each graph were defined using cosine similarity between patients' clinical feature vectors, capturing intra-age similarity in clinical profiles. Cosine similarity was selected for its computational efficiency in high-dimensional spaces, robustness to sparse data, and effectiveness in comparing embedding vectors [26]. Besides, in preliminary assessments, the clustering performance of the feature representations (evaluated using the Silhouette score) were comparable to that obtained using the Euclidean distance metric. Edges were added to connect patients with highly similar features, retaining only meaningful relationships and reducing noise. Specifically, an edge was formed when the cosine similarity between two nodes exceeded a threshold; the average 90th percentile of pairwise cosine similarity scores across all age-based graph snapshots.

### 1. Disease Stage Discovery Framework

The structured dynamic graph was fed into the dynamic graph representation model as in our previous work [16] (presented in Figure 1), which learned patient embeddings that capture both structural relationships among patients and temporal progression of their features. At each age-based graph, a graph convolutional network learns spatial embeddings that encode the local relational context of each patient within the graph by aggregating information from direct neighboring nodes (step 1 in Figure 1) [16, 27, 28]. These embeddings were subsequently organized as temporal sequences for each patient and passed through a recurrent layer to model longitudinal dependencies across visits (step 2 in Figure 1), enabling the model to capture historical and evolving clinical trajectories. The spatial and temporal representations are then fused to produce a unified latent embedding for each patient at every clinical visit (step 3 in Figure 1), capturing both feature patterns within a patient and relative to other patients, as well as the dynamics of disease progression [16, 29, 30]. In this implementation, the representation model is trained in an unsupervised manner by reconstructing node features at each graph snapshot. The model minimizes a reconstruction loss computed using mean squared error (MSE). The decoder reconstructs node attributes from learned embeddings through operations reversed the graph convolution–based construction operation, analogous to the graph decoder function in graph autoencoder frameworks [31]. During this process, a grid search conducted over different latent dimensionalities (ranging from 4 to 64 and increasing exponentially) to identify the configuration that consistently minimized the reconstruction loss across 10 random seeds. After training, the model generates latent vectors for every patient at every age transductively [32]. A transductive inference scheme is used because the objective is to characterize relationships within existed available and observed patient cohort rather than to deploy a model for prediction on unseen patients This learned representation was clustered using $K$-means++ (step 4 in Figure 1), with grid searching over different numbers of clusters ($k$) to identify the $ko$ that maximizes the silhouette score (SS) (step 4 in Figure 1). In addition, the Davies-Bouldin index (DB) and the Calinski-Harabasz index (CH) were used to evaluate the quality of the clustering [33-35].

The best $ko$ was then used as the initial clustering solution, after which, cluster stability analysis was used to investigate whether additional stable and meaningful clusters (representing disease stages) exist beyond the most parsimonious separation. Starting from the best-performing $ko$ , the number of clusters was incrementally increased, and cluster stability was assessed using bootstrap resampling (step 5 in Figure 1). Cluster assignments from bootstrap samples were compared with those obtained from the full dataset using Optimal Transport–based alignment (OTA) and the Bootstrap Jaccard coefficient (BJ) [36, 37]. Mean stability values above a threshold of 0.8, as proposed by [38], indicate robust subgroup structures. Once stability falls below this threshold the exploration stops and the last stable solution ($k$) was selected as the final number of clusters.

The resulting clusters were statistically validated using permutational multivariate analysis of variance (PERMANOVA) to assess multivariate separation [39], while differences in individual clinical measurements across the discovered clusters were examined using the Kruskal–Wallis test [40], with statistical significance set at p-value<0.05. Finally, the identified disease stages were characterized through visual analyses of clinical profiles and temporal progression patterns, contextualized with related findings from the literature.

## 3 RESULTS AND DISCUSSION

The selected subset of Enroll-HD data utilized in this study comprised longitudinal records from 302 patients, with 1,028 clinical visits and 44 clinical features. As shown in Table 1, most participants were in the manifest stage (80.5%). The majority of the participants were aged in their 50s, and most had a CAG repeat length of 42. Patients' clinical profiles represented by URL-STFN consistently achieved the best clustering performance with compact latent dimensions of four (4 for spatial, 4 for temporal, and 4 for fusion) and produced consistent results across the tested seeds. The generated clinical visits embeddings were clustered, $K$-means++ reached its highest clustering performance with a two-cluster solution, achieving mean ± standard deviation scores of 0.670 ± 0.027, 0.568 ± 0.037, and 453.98 ± 75.25 for the SS, DB, and CH, respectively. When these representations were used in stability assessments to identify additional clusters, clustering remained robust up to four clusters, which were found to be significantly distinct according to the PERMANOVA test (p-value=0.001). However, inter-cluster separation and intra-cluster compactness decreased slightly, with SS = 0.51, DB = 0.66, and CH = 1528.97.

Table 1: Clinical characteristics of the Enroll-HD dataset subset used for uncovering HD disease stages (main dataset with four longitudinal visits with common ages).

| Dataset | Number of participants | Number of observations | Age range | CAG repeats (mean ± SD) | Premanifest (count,%) | Manifest (count,%) |
|---|---|---|---|---|---|---|
| Main Dataset | 302 | 1208 | 54.0–57.0 | 42.5 ± 1.9 | 234 (19.4%) | 972 (80.5%) |

### 3.1 Disease Stage Characteristics

The interesting aspect of the clustering results lies in the clinical characteristics revealed by each discovered cluster. To explore how these clusters relate to the clinical progression of HD, we first visualized the cluster assignments across patients' longitudinal visits using heatmaps. Specifically, patient records were organized according to the UHDRS DCL, where visits categorized as premanifest (DCL < 4) and manifest (DCL = 4) are plotted in two separate heatmaps (see Figure 2). This stratification provides an abstract view of how the identified stages are positioned along the disease continuum, with the support of established HD clinical scales. Cluster 0 was predominantly represented during the premanifest phase, suggesting that it captures early or subtle disease-related changes prior to formal motor diagnosis. Cluster 1 appeared in both premanifest and manifest visits but was observed more frequently after manifest conversion, indicating a potential transitional or early manifest phenotype. In contrast, clusters 2 and 3 were largely confined to manifest visits, suggesting that they represent later or more advanced disease stages.

The severity gradients of the discovered clusters become more apparent when visualizing the mean standardized values of the clinical features in each cluster. This pattern is illustrated by the hue-colored heatmap (Figure 3). This visualization demonstrates how patients' clinical status progressively worsens across clusters: cluster 0 represents a relatively preserved clinical state, whereas clinical features gradually deteriorate, reaching the most severe impairment in cluster 3.

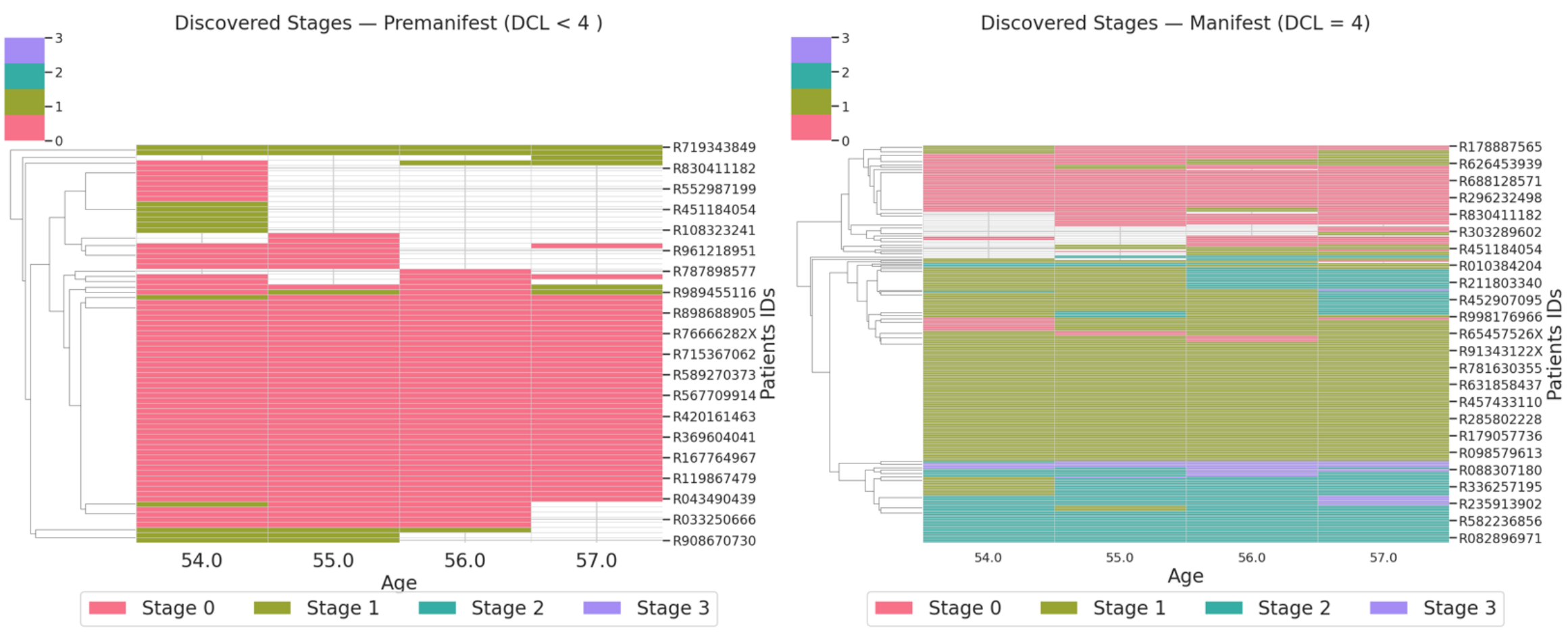

Figure 2: Heatmaps show the distribution of discovered clusters across patient visits. The clinical premanifest heatmap (DCL < 4) displays visits occurring in the premanifest stage, while the manifest panel (DCL = 4) shows visits classified as manifest, with colors indicating the corresponding cluster assignment. Rows represent subjects and are hierarchically clustered to highlight similarities in longitudinal trajectories; columns represent common ages at visit. Colors correspond to actual cluster numbers (legend), with white cells indicate patients visits that do not belong to the current stage depicted in the heatmap.

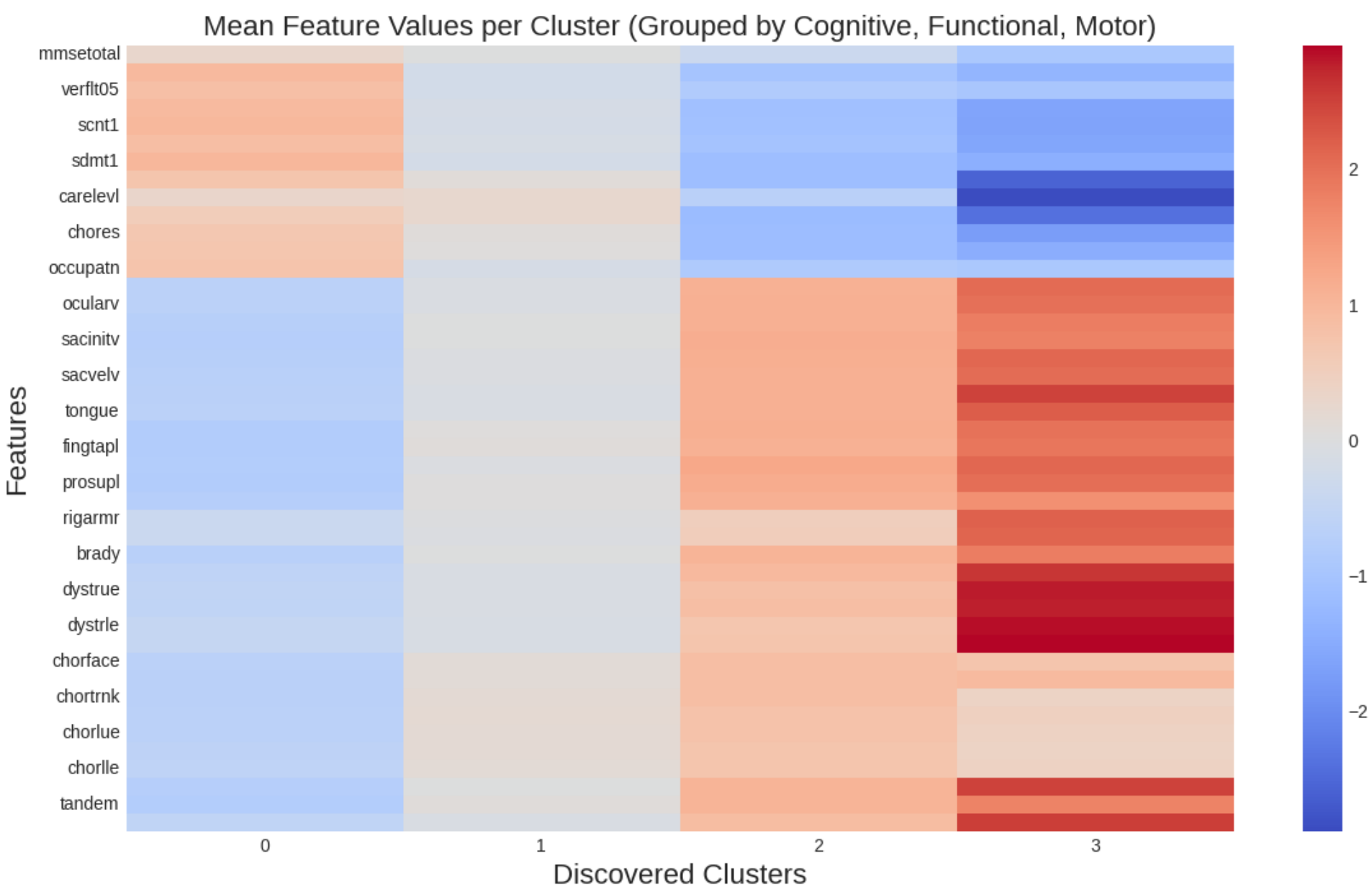

Figure 3: Heatmap shows the mean values of all features across the discovered clusters. Features in Y-axis are grouped by domain: cognitive (top), functional (middle), and motor (bottom). This visualization highlights domain-specific patterns and the variation of feature severity across clusters.

All features exhibited significant differences across the discovered clusters, with p-values<0.05 as determined by the Kruskal–Wallis H-test, with Bonferroni correction applied to adjust for multiple comparisons across features. To characterize the clinical profiles of the discovered clusters and provide quantitative support for the observed clinical measurements patterns, we generated violin plots to illustrate the distribution of individual features across clusters (Appendix A4; Figures A1 to A3). In addition, the mean values of these features and their 95% confidence interval (CI) across the discovered clusters are reported in Table A1 in the appendix. Notably, both data presentations demonstrate how patients' cognitive, functional, and motor measurements degraded progressively from cluster 0 to cluster 3. The decline in the cognitive measures is apparent in Figure A1. For instance, mini-mental state examination (MMSE) mean scores decreased from 28.53 in cluster 0 to 25.49 in cluster 1, 19.73 in cluster 2, and 18.70 in cluster 3 (see Table A1). A similar pattern was observed for the SDMT, with mean [95%CI] scores of 44.11 [43.00–45.21], 23.14 [22.39–23.90], 9.69 [8.55–10.83], and 6.33 [-1.58–14.25] across clusters 0–3, respectively. Likewise, Stroop Word Reading (swrt1) scores declined from 88.46 [86.58–90.34] in cluster 0 to 54.91 [53.35–56.46], 29.87 [27.85–31.89], and 16.38 [8.68–24.07] in clusters 1–3. Similarly, the decline trend continue with the reset cognitive features, including Verbal Fluency (verflt05) and Stroop Color Naming (scnt1) scores.

Moreover, functional capacity patterns across clusters (presented in Figure A2) demonstrated one of the steepest gradients. Independence scale scores declined from 93.18 [92.27–94.09] in cluster 0 to 80.61 [79.60–81.62], 56.00 [54.35–57.65], and 27.88 [22.60–33.15] in clusters 1–3, reflecting the need for increased care over time. In addition, activities of daily living (adl) scores decreased from 2.90 [2.86–2.93] to 2.62 [2.57–2.66], 1.34 [1.23–1.45], and 0.30 [0.12–0.48] across the clusters. Clinically, this transition is highly relevant for prognosis, caregiver burden, and resource planning, as it likely marks the stage at which patients move from partial independence to sustained dependency.

In parallel, motor impairment increased steadily across clusters, as depicted in Figure A3. Bradykinesia (brady) scores rose from 0.38 [0.32–0.44] in cluster 0 to 1.24 [1.15–1.32], 2.48 [2.35–2.61], and 3.45 [3.12–3.79] in clusters 1–3, respectively, while gait scores increased from 0.27 [0.22–0.32] to 1.01 [0.95–1.07], 2.03 [1.94–2.11], and 3.45 [3.18–3.73]. Postural instability and tandem gait impairment became particularly pronounced in later clusters, contributing substantially to functional decline. Similarly, oculomotor abnormalities followed the same pattern, with vertical saccade velocity (sacvelv) increasing from 0.52 [0.46–0.58] in cluster 0 to 1.29 [1.22–1.37], 2.64 [2.50–2.77], and 3.73 [3.51–3.94] in cluster 3. Such oculomotor abnormalities–including impaired saccadic velocity and initiation–are known to emerge early and worsen progressively [30]. Another noteworthy chorea-related observation is the deterioration of the hyperkinetic choreiform movements (chorlle, chorrle, chorlue, chorrue, . . . ) where severity increased from early to intermediate stages but appeared to attenuate in cluster 3 (see the top bar plot in Figure 4). This pattern aligns with clinical observations, where chorea often emerges early and intensifies during the initial manifest stages, whereas in later stages it levels off and is replaced or accompanied by bradykinesia (brady), rigidity (rigarmr, rigarml), and dystonia (dysttrnk, dystrue, . . . ) [42].

The consistency of these patterns across multiple measurement domains–such as motor sub-scores (rigidity, bradyki-nesia, dystonia, gait), cognitive assessments, and functional indices–reinforces the internal meaningful coherence of the clustering solution. This suggests that the discovered stages are not defined by isolated features but by coordinated, multi-domain measurement changes, supporting the hypothesis that the clusters identified by the proposed framework represent clinically meaningful disease phenotypes. While clustering and stability-assessment methods can sometimes create apparent differences due to the algorithms used to process the data, the consistent multi-domain patterns indicate that the clusters reflect true disease progression rather than just random grouping artifacts..

### 3.2 Mapping Discovered Stages to Established Clinical Staging Systems

The discovered URL-STFN-based clusters were systematically compared with the established HD staging frameworks: the DCL-based motor diagnosis within the UHDRS, the Shoulson and Fahn functional staging system, and HD-ISS [7]. Given the calculated mean UHDRS TMS for cluster 0 was 13.75 obtained by summing the mean motor feature scores (Table A1) [27]. In large longitudinal cohorts such as PREDICT-HD and TRACK-HD reported average TMS values closer to 15 at the time of confirmed motor diagnosis (DCL = 4), although values can vary widely between individuals, ranging roughly from 8 to 35 [44]. Besides, in Long et al. [43] considered 13.75 part of a manifest stage (DCL=4), suggesting that the discovered cluster 0 may correspond more to a phenoconversion stage.

For the discovered stages 1, 2, and 3, cluster 1 appears to represent a transitional phase that can lie between Shoulson and Fahn stages II and III, while reported TFC in the discovered stages 2 and 3 fall within Shoulson and Fan TFC stage III range. This suggests that our framework may divide Shoulson and Fahn Stages III into finer sub-stages in a meaningful way, with each sub-stage reflecting distinct patterns in TFC and other clinical measurements rather than arising by chance.

The distributions of clinical features were summarized using the median and interquartile range (Q1–Q3) for each variable across all staging systems and visualized as line graphs in Figure 4, allowing direct comparison of the central tendency and clinical measurements across the discovered stages and the established stages of existing HD staging systems, while accounting for their dispersion and overlap.

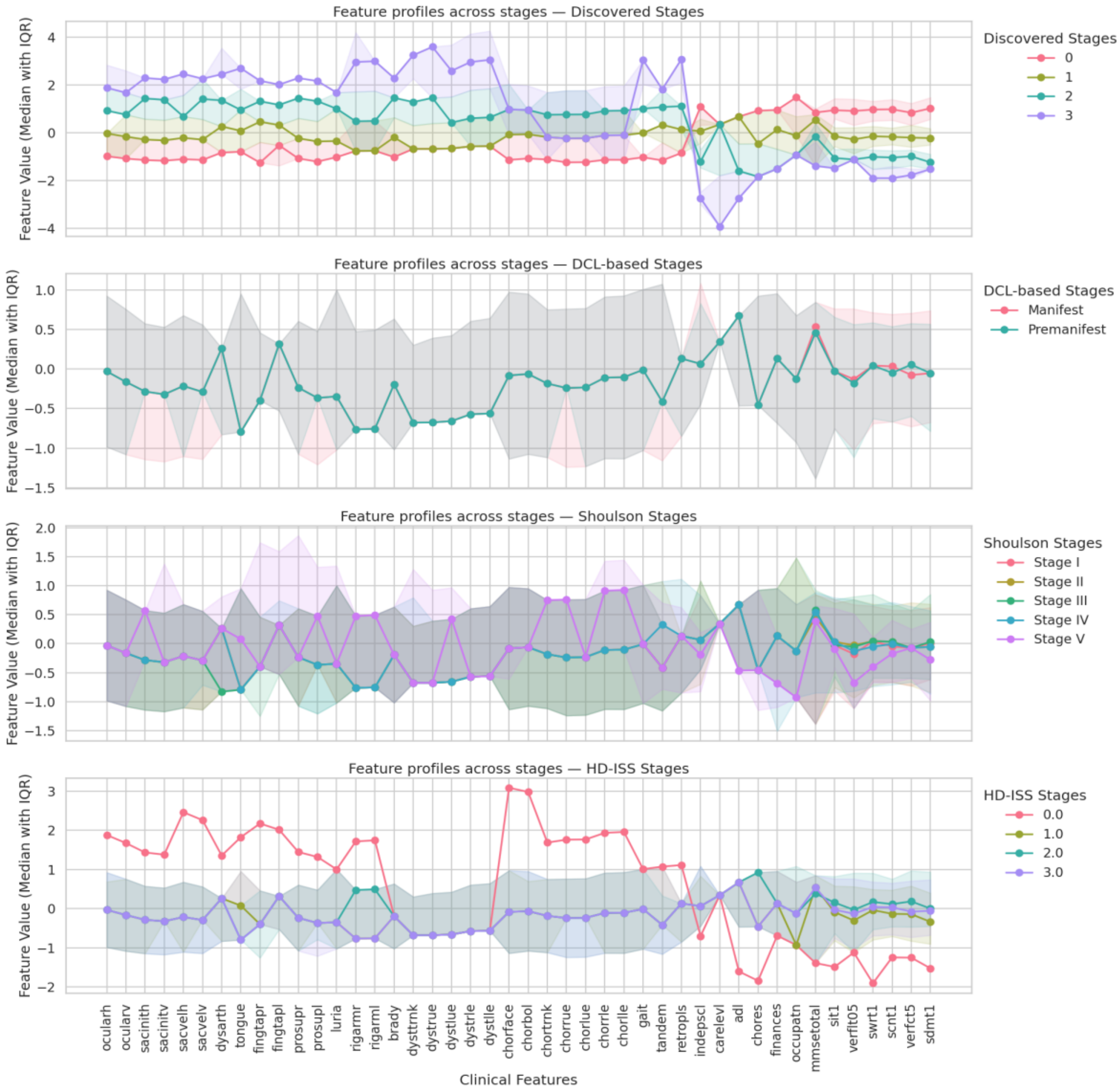

Figure 4: Line plots for the standardized median clinical feature values across the proposed framework discovered stages (Top graph), and conventional HD staging systems named in the figure as HD-ISS, Shoulson and Fahn, and DCL-based stages (ordered from bottom to top plot, respectively).

As hypothesized, the stages identified by the proposed framework demonstrated the clearest separation. Their progress were characterized by steady shifts in median values and minimal interquartile overlap between adjacent stages. In particular, cognitive measures showed clear stepwise declines with limited overlap between early and late clusters. Functional measures followed a similar pattern, providing strong discrimination between stages. Motor variables, including bradykinesia, gait performance, tandem walking, and saccadic movements, further strengthened stage differentiation, each showing minimal median overlap between clusters. Notable, previous studies have reported limited inter-rater reliability for several motor elements of UHDRS, particularly tongue protrusion (tongue), maximal trunk dystonia (dysttrnk) and arm rigidity (rigarml), indicating that these features can be difficult to rate consistently in clinical examinations [9]. However, the discovered stages demonstrated the ability to discriminate meaningful ranges of these features across the discovered disease stages, as demonstrated in Figure 4 that shows the clear progression of these features with minimal overlap.

To detail, tongue protrusion (tongue) shows a clear progression across the discovered clusters, with median [IQR range] increasing from 0.00 [0.00–0.00] in cluster 0 to 1.00 [0.00–2.00] in cluster 1, 2.00 [2.00–3.00] in cluster 2, and 4.00 [3.00–4.00] in cluster 3, indicating a strong separation between stages. A similar pattern is observed for trunk dystonia (dysttrnk), which progresses from 0.00 [0.00–0.00] in cluster 0 to 0.00 [0.00–1.00] in cluster 1, 2.00 [1.00–2.00] in cluster 2, and 4.00 [3.00–4.00] in cluster 3. Likewise, left arm rigidity (rigarml) demonstrates gradual stage differentiation.

Minimal features dispersion overlap is observed for the discovered stages. Early stages features were confined more than later stages for all features, except for rigidity and chorea related measurements. MMSE cognitive feature not counted since its missing rate was 48% and imputed with -1 (require further investigation), suggesting that the proposed ML-driven staging framework captures progressive motor deterioration more distinctly than conventional staging systems, which often rely on broader categorical thresholds. In contrast, DCL-based groupings showed broader interquartile ranges and greater dispersion between premanifest and manifest categories. Similarly, HD-ISS and Shoulson and Fahn staging systems provided less granularity when capturing transitions from early to advanced disease stages.

Taken together, multidomain clinical markers—particularly processing speed, executive function, bradykinesia, gait impairment, oculomotor slowing, and functional independence—emerged as the main contributors to stage differentiation. Within this comparison, the framework's clusters provided the most internally consistent stratification, with clearer separation and reduced overlap across severity levels. However, Since this conclusion is based on data-driven patterns rather than clinical evaluation, it should still be considered hypothetical and subject to clinical expert validation.

## 4 CONCLUSION

Although the framework was evaluated on a limited subset of the Enroll-HD dataset, the discovered stages distinguished multidomain clinical features more clearly than conventional staging systems. This observation motivates further analysis using the full dataset to investigate whether additional disease stages can be identified. Stages separations were pronounced, despite the model being derived in an unsupervised manner without ground truth. These findings indicate that the employment of the presented URL-STFN-based framework's component has effectively encoded clinically meaningful severity patterns, enabling the discovery of distinct disease stages. The framework not only captures these inter- and intra- stages based TFC scores but also maps features scores with minimal overlap between severity levels, all achieved without relying on explicit clinical supervision.

Clinical expert validation of the clinical profiles of the discovered stages will be an important focus of future work. Some clinical measurements remain subjective and dependent on expert assessment and may vary across raters [9]. Therefore, analyzing morphometric and/or biofluids biomarker remains an unmet need and represents a potential avenue for future work to uncover purely objective disease states. Future work will also focus on explaining the model decision that derive the latents construction and the discovery for the four clusters, enabling a clearer identification of key motors, functional and cognitive biomarkers driving cluster transition and separation. Furthermore, the missingness imputation with -1 is a simplified strategy but was sufficient for the proof-of-concept purpose. However, future extensions will explore masking-based approaches to handle missingness more rigorously, thereby improving the reliability of the learned representations and, consequently, the quality of the clusters discovered.

Overall, the proposed unsupervised framework provides a data-driven, multidomain staging approach for HD that complements existing clinical scales, captures disease severity gradients, and encourages ongoing refinement and validation to establish it as a clinically validated ML-based staging system.

## ACKNOWLEDGMENTS

Data used in this work was generously provided by the participants in the Enroll-HD study and made available by CHDI Foundation, Inc. Enroll-HD is a global clinical research platform intended to accelerate progress towards therapeutics for Huntington’s disease; core datasets are collected annually on all research participants as part of this multi-center longitudinal observational study. Enroll-HD is sponsored by CHDI Foundation, Inc., a nonprofit biomedical research organization exclusively dedicated to developing therapeutics for Huntington’s disease. Enroll-HD would not be possible without the vital contribution of the research participants and their families [18].

## A APPENDICES

### A.1 Data Availability

The Enroll-HD dataset used in the assessment of the proposed graph model can be requested directly from Enroll-HD clinical research platform; https://www.enroll-hd.org/. For code access and work reproducibility, contact the corresponding author.

### A.2 Conflict of interest

The authors have no conflicts of interest to declare that are relevant to the content of this article.

### A.3 Data Availability

Generative AI was used only for language polishing and formatting. All scholarly content is the sole responsibility of the authors.

### A.4 Additional Figures & Tables

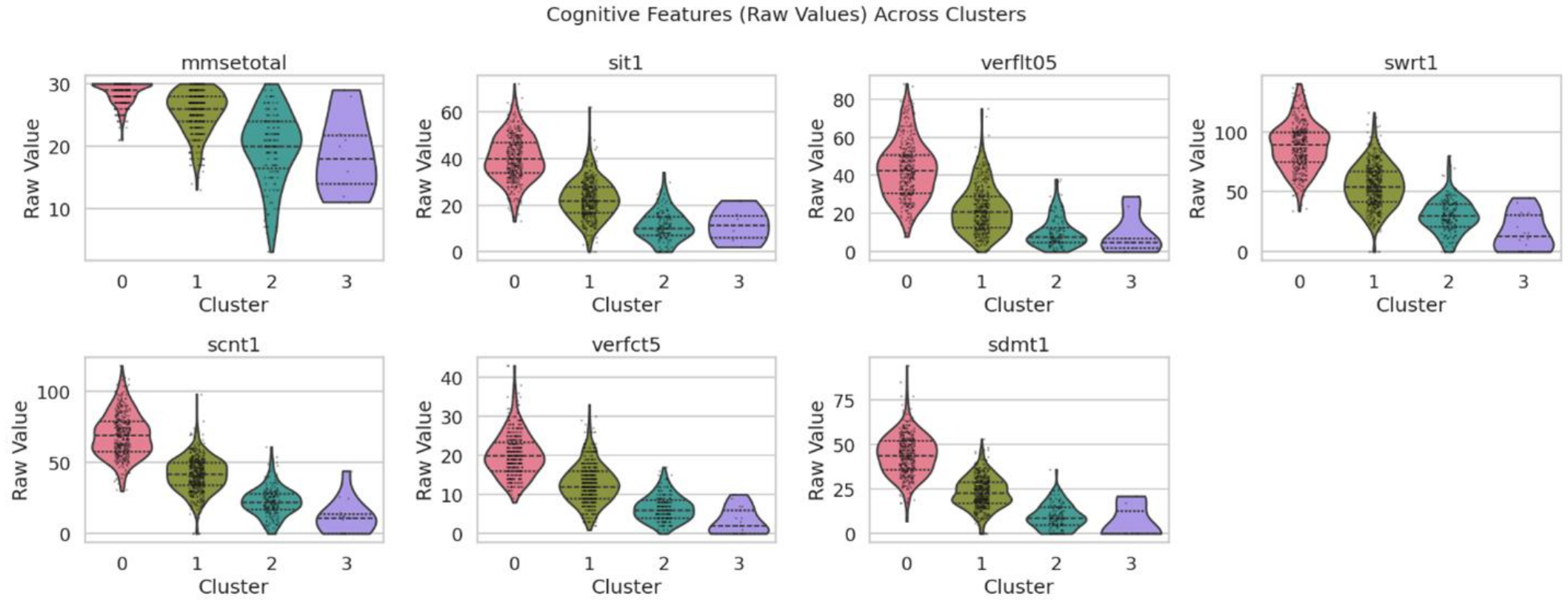

Figure A1: Violin plots of cognitive features across the URL-STFN based discovered stages (clusters 0–3) in Huntington's disease. Displayed measures were statistically significant across stages, and their distributions demonstrate progressive functional decline with advancing discovered stage.

Table A1: Feature means for URL-STFN-based framework discovered stages with grouped summed feature scores for every feature type and references.

| Feature | Cluster 0 | Cluster 1 | Cluster 2 | Cluster 3 |
|---|---|---|---|---|
| mmsetotal | 28.53 [28.33–28.73] | 25.49 [25.14–25.84] | 19.73 [18.78–20.68] | 18.70 [14.75–22.65] |
| verflt05 | 42.85 [41.33–44.37] | 22.33 [21.13–23.52] | 10.21 [8.77–11.66] | 8.78 [1.94–15.61] |
| sit1 | 40.28 [39.34–41.23] | 22.60 [21.84–23.36] | 11.26 [10.17–12.36] | 11.33 [5.40–17.27] |
| sdmt1 | 44.11 [43.00–45.21] | 23.14 [22.39–23.90] | 9.69 [8.55–10.83] | 6.33 [-1.58–14.25] |
| swrt1 | 88.46 [86.58–90.34] | 54.91 [53.35–56.46] | 29.87 [27.85–31.89] | 16.38 [8.68–24.07] |
| scnt1 | 69.50 [68.01–70.98] | 41.40 [40.36–42.44] | 22.46 [21.00–23.91] | 11.93 [5.70–18.17] |
| verfct5 | 20.25 [19.70–20.79] | 12.81 [12.38–13.24] | 6.60 [6.13–7.07] | 3.24 [1.61–4.86] |
| rigarmr | 0.30 [0.26–0.35] | 0.60 [0.54–0.65] | 1.01 [0.89–1.13] | 2.36 [1.86–2.87] |
| retropls | 0.29 [0.24–0.34] | 0.78 [0.71–0.84] | 1.79 [1.66–1.91] | 3.45 [3.16–3.75] |
| rigarml | 0.31 [0.26–0.36] | 0.56 [0.51–0.62] | 1.02 [0.90–1.14] | 2.30 [1.82–2.79] |
| sacinitv | 0.50 [0.44–0.56] | 1.37 [1.29–1.44] | 2.73 [2.62–2.84] | 3.48 [3.20–3.77] |
| chorlle | 0.53 [0.47–0.60] | 1.23 [1.16–1.30] | 1.79 [1.66–1.92] | 1.48 [1.02–1.95] |
| chorrle | 0.50 [0.44–0.57] | 1.26 [1.18–1.33] | 1.82 [1.69–1.95] | 1.48 [1.02–1.95] |
| chorlue | 0.58 [0.52–0.65] | 1.39 [1.32–1.46] | 2.01 [1.88–2.14] | 1.64 [1.17–2.10] |
| prosupl | 0.47 [0.41–0.53] | 1.45 [1.38–1.52] | 2.83 [2.72–2.94] | 3.82 [3.68–3.95] |
| fingtapr | 0.51 [0.45–0.57] | 1.51 [1.44–1.58] | 2.79 [2.67–2.90] | 3.76 [3.59–3.93] |
| tongue | 0.17 [0.12–0.21] | 0.81 [0.73–0.88] | 2.16 [2.03–2.30] | 3.45 [3.20–3.71] |
| fingtapl | 0.67 [0.60–0.74] | 1.69 [1.62–1.76] | 2.91 [2.80–3.01] | 3.88 [3.77–3.99] |
| prosupr | 0.35 [0.30–0.40] | 1.23 [1.16–1.29] | 2.76 [2.65–2.88] | 3.79 [3.62–3.95] |
| sacvelh | 0.44 [0.38–0.50] | 1.19 [1.13–1.26] | 2.50 [2.37–2.62] | 3.61 [3.38–3.83] |
| sacvelv | 0.52 [0.46–0.58] | 1.29 [1.22–1.37] | 2.64 [2.50–2.77] | 3.73 [3.51–3.94] |
| sacinith | 0.49 [0.42–0.56] | 1.32 [1.24–1.39] | 2.62 [2.51–2.73] | 3.45 [3.16–3.75] |
| luria | 0.41 [0.34–0.48] | 1.54 [1.43–1.65] | 3.16 [3.03–3.30] | 3.85 [3.66–4.04] |
| dysttrnk | 0.10 [0.07–0.14] | 0.59 [0.52–0.65] | 1.65 [1.50–1.79] | 3.33 [3.07–3.60] |
| dystrue | 0.13 [0.09–0.16] | 0.55 [0.49–0.61] | 1.39 [1.26–1.52] | 3.24 [2.92–3.56] |
| dystlue | 0.10 [0.07–0.13] | 0.52 [0.46–0.58] | 1.40 [1.27–1.53] | 3.15 [2.83–3.47] |
| brady | 0.38 [0.32–0.44] | 1.24 [1.15–1.32] | 2.48 [2.35–2.61] | 3.45 [3.12–3.79] |
| dystrle | 0.09 [0.06–0.13] | 0.39 [0.33–0.45] | 1.09 [0.96–1.21] | 2.91 [2.59–3.23] |
| finances | 2.71 [2.65–2.78] | 1.88 [1.79–1.97] | 0.40 [0.31–0.49] | 0.03 [-0.03–0.09] |
| chorrue | 0.58 [0.51–0.65] | 1.41 [1.33–1.48] | 2.01 [1.88–2.13] | 1.70 [1.24–2.15] |
| chortrnk | 0.48 [0.42–0.55] | 1.35 [1.27–1.43] | 2.10 [1.96–2.24] | 1.61 [1.15–2.06] |
| chorbol | 0.40 [0.35–0.46] | 1.17 [1.10–1.24] | 1.90 [1.77–2.02] | 1.97 [1.48–2.46] |
| chorface | 0.46 [0.39–0.52] | 1.18 [1.12–1.25] | 1.89 [1.77–2.01] | 1.76 [1.32–2.19] |
| dysarth | 0.13 [0.10–0.17] | 0.68 [0.62–0.73] | 1.78 [1.67–1.89] | 3.03 [2.73–3.33] |
| ocularv | 0.46 [0.40–0.52] | 1.10 [1.03–1.17] | 2.38 [2.26–2.49] | 3.33 [3.11–3.55] |
| ocularh | 0.34 [0.29–0.39] | 0.96 [0.89–1.03] | 2.18 [2.06–2.29] | 3.18 [2.89–3.47] |
| dystlle | 0.08 [0.05–0.11] | 0.37 [0.31–0.42] | 1.07 [0.95–1.19] | 2.88 [2.54–3.22] |
| gait | 0.27 [0.22–0.32] | 1.01 [0.95–1.07] | 2.03 [1.94–2.11] | 3.45 [3.18–3.73] |
| tandem | 0.51 [0.43–0.58] | 1.64 [1.55–1.73] | 2.96 [2.84–3.09] | 3.94 [3.86–4.02] |
| adl | 2.90 [2.86–2.93] | 2.62 [2.57–2.66] | 1.34 [1.23–1.45] | 0.30 [0.12–0.48] |
| occupatn | 2.11 [2.00–2.22] | 0.95 [0.86–1.04] | 0.07 [0.04–0.11] | 0.00 [0.00–0.00] |
| chores | 1.84 [1.80–1.87] | 1.38 [1.34–1.43] | 0.47 [0.40–0.54] | 0.06 [-0.02–0.14] |
| indepscl | 93.18 [92.27–94.09] | 80.61 [79.60–81.62] | 56.00 [54.35–57.65] | 27.88 [22.60–33.15] |
| carelevl | 1.99 [1.97–2.00] | 1.95 [1.94–1.97] | 1.52 [1.43–1.60] | 0.48 [0.26–0.71] |

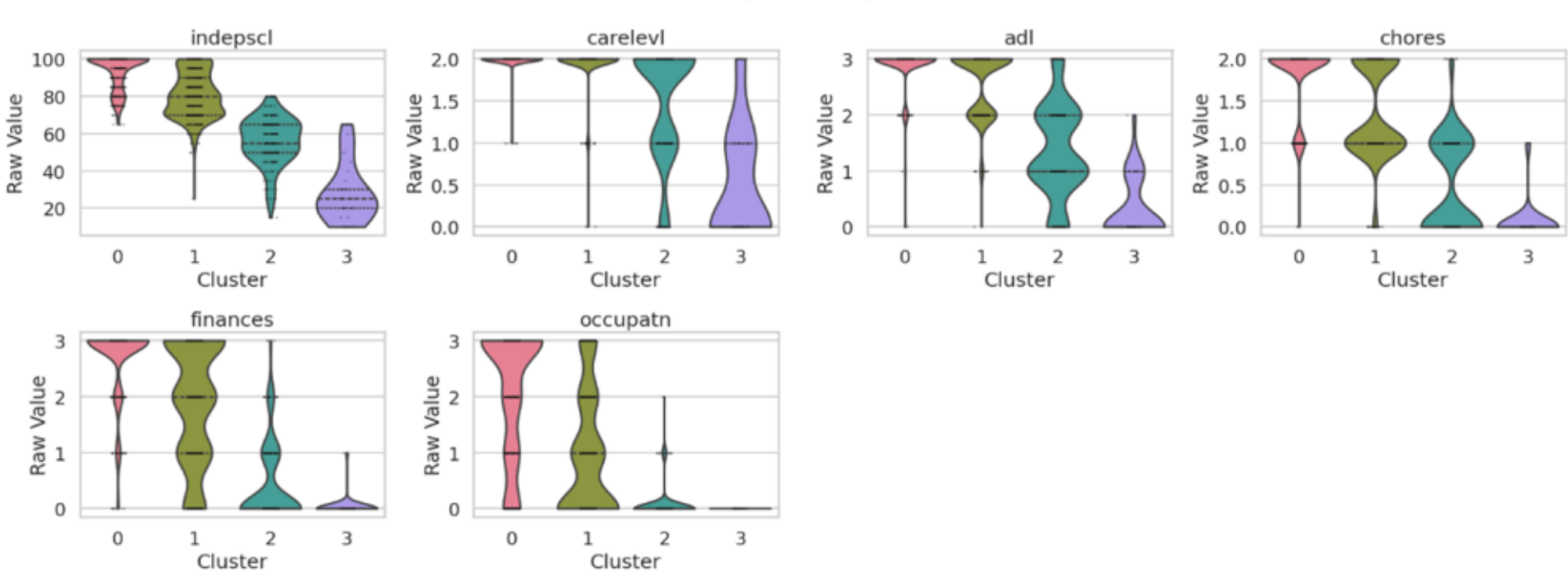

Figure A2. Violin plots of functional features across the URL-STFN based discovered stages (clusters 0–3) in Huntington's disease. Displayed measures were statistically significant across stages, and their distributions demonstrate progressive cognitive decline with advancing discovered stage.

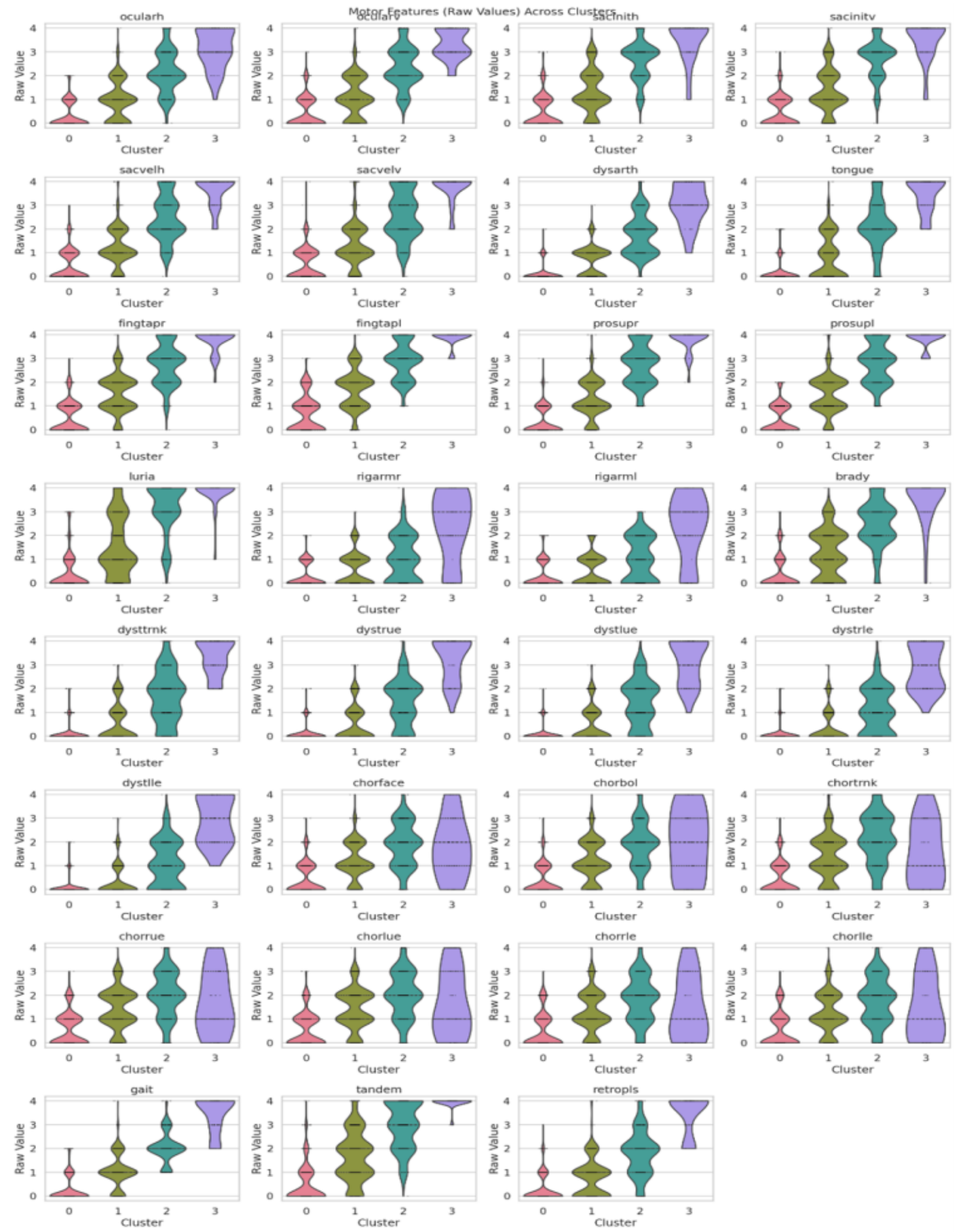

Fig. A2. Violin plots of functional features across the URL-STFN based discovered stages (clusters 0–3) in Huntington's disease. Displayed measures were statistically significant across stages, and their distributions demonstrate progressive cognitive decline with advancing discovered stage.